%% file: main.tex
\documentclass[runningheads,a4paper]{llncs}
\usepackage{booktabs}
\usepackage{multirow}
\usepackage{amssymb}
\usepackage{graphicx}
\usepackage{hyperref}
\usepackage{cleveref}
\usepackage{booktabs}
\usepackage{amsmath}
\usepackage{url}
\urldef{\mailsa}\path|{alfred.hofmann, ursula.barth, ingrid.haas, frank.holzwarth,|
\urldef{\mailsb}\path|anna.kramer, leonie.kunz, christine.reiss, nicole.sator,|
\urldef{\mailsc}\path|erika.siebert-cole, peter.strasser, lncs}@springer.com|    
\newcommand{\keywords}[1]{\par\addvspace\baselineskip
\noindent\keywordname\enspace\ignorespaces#1}
\usepackage{todonotes}
\let\llncssubparagraph\subparagraph
\let\subparagraph\paragraph
\usepackage[compact]{titlesec}
\let\subparagraph\llncssubparagraph

\titleformat{\chapter}[display]
{\normalfont\huge\bfseries}{\chaptertitlename\ \thechapter}{20pt}{\Huge}
\titleformat{\section}
{\normalfont\Large\bfseries}{\thesection}{1em}{}
\titleformat{\subsection}
{\normalfont\large\bfseries}{\thesubsection}{1em}{}
\titleformat{\subsubsection}
{\normalfont\normalsize\bfseries}{\thesubsubsection}{1em}{}
\titleformat{\paragraph}[runin]
{\normalfont\normalsize\bfseries}{\theparagraph}{1em}{}
\titleformat{\subparagraph}[runin]
{\normalfont\normalsize\bfseries}{\thesubparagraph}{1em}{}

\titlespacing*{\chapter} {0pt}{50pt}{40pt}
\titlespacing*{\section} {0pt}{3.5ex plus 1ex minus .2ex}{2.3ex plus .2ex}
\titlespacing*{\subsection} {0pt}{3.25ex plus 1ex minus .2ex}{1.5ex plus .2ex}
\titlespacing*{\subsubsection}{0pt}{3.25ex plus 1ex minus .2ex}{1.5ex plus .2ex}
\titlespacing*{\paragraph} {0pt}{3.25ex plus 1ex minus .2ex}{1em}
\titlespacing*{\subparagraph} {\parindent}{3.25ex plus 1ex minus .2ex}{1em}
\DeclareMathOperator*{\argmin}{arg\,min}

\begin{document}

\mainmatter  

\title{Automatic Liver and Lesion Segmentation in CT Using Cascaded Fully Convolutional Neural Networks and 3D Conditional Random Fields}


%
%
\author{Patrick Ferdinand Christ\inst{1} \and Mohamed Ezzeldin A. Elshaer\inst{1} \and Florian Ettlinger\inst{1} \and Sunil Tatavarty\inst{2} \and Marc Bickel\inst{1} \and  Patrick Bilic\inst{1} \and   Markus Rempfler\inst{1} \and Marco Armbruster\inst{4} \and Felix Hofmann\inst{4} \and Melvin D'Anastasi\inst{4} \and Wieland H. Sommer\inst{4} Seyed-Ahmad Ahmadi\inst{3} \and Bjoern H. Menze\inst{1}}

\institute{Technische Universit\"at M\"unchen, Image-Based Biomedical Modeling Group, Arccisstrasse 21, 80333 Munich, Germany\\
Patrick.Christ@tum.de and Bjoern.Menze@tum.de
\and 
Technische Universit\"at M\"unchen, Chair for Data Processing,
Arccisstrasse 21, 80333 Munich, Germany
\and
LMU Hospital Grosshadern, Department for Neurology, Marchioninistrasse 15, 81377 Munich, Germany \and
LMU Hospital Grosshadern, Department for Clinical Radiology, Marchioninistrasse 15, 81377 Munich, Germany}
\authorrunning{Christ et al.}
\titlerunning{Automatic Liver and Lesion Segmentation in CT using CFCNs and 3DCRFs}



%

\toctitle{Lecture Notes in Computer Science}
\tocauthor{Authors' Instructions}
\maketitle

\begin{abstract}
Automatic segmentation of the liver and its lesion is an important step towards deriving quantitative biomarkers for accurate clinical diagnosis and computer-aided decision support systems. This paper presents a method to automatically segment liver and lesions in CT abdomen images using cascaded fully convolutional neural networks (CFCNs) and dense 3D conditional random fields (CRFs).  
We train and cascade two FCNs for a combined segmentation of the liver and its lesions. In the first step, we train a FCN to segment the liver as ROI input for a second FCN. The second FCN solely segments lesions from the predicted liver ROIs of step 1. We refine the segmentations of the CFCN using a dense 3D CRF that accounts for both spatial coherence and appearance.
CFCN models were trained in a 2-fold cross-validation on the abdominal CT dataset 3DIRCAD comprising 15 hepatic tumor volumes. Our results show that CFCN-based semantic liver and lesion segmentation achieves Dice scores over $94\%$ for liver with computation times below 100s per volume. We experimentally demonstrate the robustness of the proposed method as a decision support system with a high accuracy and speed for usage in daily clinical routine.
\end{abstract}

\input{content}
\bibliographystyle{splncs03}
\bibliography{literature.bib}

\end{document}

%% file: content.tex

\newcommand{\labelspace}{\mathcal{\mathcal{L}}}     
\newcommand{\x}{\mathbf{x}} 
\newcommand{\vertices}{\mathcal{V}} 
\newcommand{\edges}{\mathcal{E}}    
\newcommand{\graph}{\mathcal{G}}    
\newcommand{\cp}[2]{P\left( #1 \vert #2 \right)}    

\section{Introduction}

Anomalies in the shape and texture of the liver and visible lesions in CT are important biomarkers for disease progression in primary and secondary hepatic tumor disease \cite{Heimann}.
In clinical routine, manual or semi-manual techniques are applied. These, however, are subjective, operator-dependent and very time-consuming. In order to improve the productivity of radiologists, computer-aided methods have been developed in the past, but the challenges in automatic segmentation of combined liver and lesion remain, such as low-contrast between liver and lesion, different types of contrast levels (hyper-/hypo-intense tumors), abnormalities in tissues (metastasectomie), size and varying amount of lesions.

Nevertheless, several interactive and automatic methods have been developed to segment the liver and liver lesions in CT volumes. In 2007 and 2008, two Grand Challenges benchmarks on liver and liver lesion segmentation have been conducted \cite{Heimann,deng2008editorial}. Methods presented at the challenges were mostly based on statistical shape models. Furthermore, grey level and texture based methods have been developed \cite{Heimann}. Recent work on liver and lesion segmentation employs graph cut and level set techniques \cite{li2015automatic,li2013likelihood,linguraru2012tumor}, sigmoid edge modeling \cite{foruzan2015improved} or manifold and machine learning \cite{kadoury2015metastatic,freiman2011liver}. However, these methods are not widely applied in clinics, due to their speed and robustness on heterogeneous, low-contrast real-life CT data. Hence, interactive methods were still developed \cite{hame2012semi,ben2015automated} to overcome these weaknesses, which yet involve user interaction.

Deep Convolutional Neural Networks CNN have gained new attention in the scientific community for solving computer vision tasks such as object recognition, classification and segmentation \cite{krizhevsky2012imagenet,long2014fully}, often out-competing state-of-the art methods. Most importantly, CNN methods have proven to be highly robust to varying image appearance, which motivates us to apply them to fully automatic liver and lesions segmentation in CT volumes. 

Semantic image segmentation methods based on fully convolutional neural networks FCN were developed in \cite{long2014fully}, with impressive results in natural image segmentation competitions \cite{chen2014semantic,zheng2015conditional}. Likewise, new segmentation methods based on CNN and FCNs were developed for medical image analysis, with highly competitive results compared to state-of-the-art. \cite{Unet,Brats,wang2015detection,roth2015deeporgan,prasoon13,kamnitsas2016efficient}.

In this work, we demonstrate the combined automatic segmentation of the liver and its lesions in low-contrast heterogeneous CT volumes. Our contributions are three-fold. First, we train and apply fully convolutional CNN on CT volumes of the liver for the first time, demonstrating the adaptability to challenging segmentation of hepatic liver lesions. Second, we propose to use a cascaded fully convolutional neural network (CFCN) on CT slices, which segments liver and lesions sequentially, leading to significantly higher segmentation quality. Third, we propose to combine the cascaded CNN in 2D with a 3D dense conditional random field approach (3DCRF) as a post-processing step, to achieve higher segmentation accuracy while preserving low computational cost and memory consumption. In the following sections, we will describe our proposed pipeline (\Cref{sec:pipeline}) including CFCN (\Cref{sec:cfcn}) and 3D CRF (\Cref{sec:3dcrf}), illustrate experiments on the 3DIRCADb dataset (\Cref{sec:results}) and summarize the results (\Cref{sec:summary}).

\section{Methods}

\label{sec:results}
\begin{figure*}[tbh]
\centering
\includegraphics[width=\columnwidth]{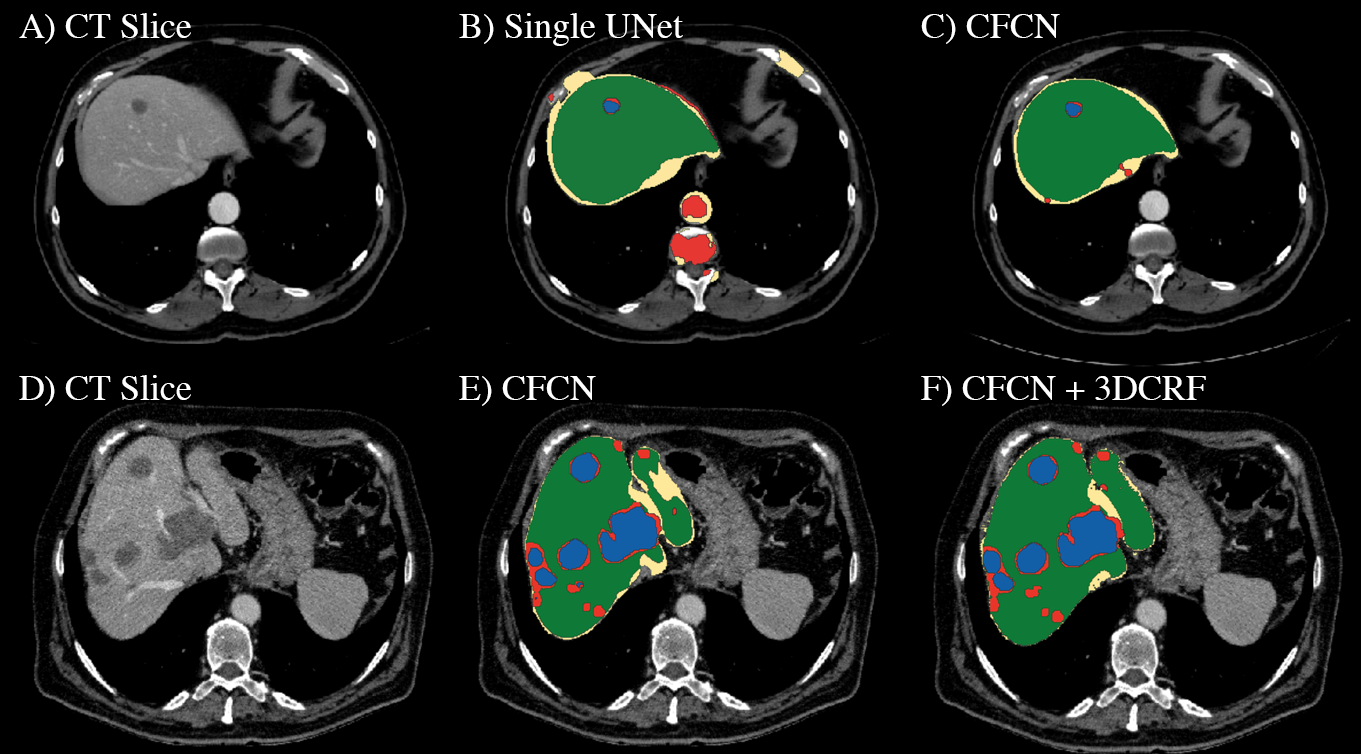}
\caption{Automatic liver and lesion segmentation with cascaded fully convolutional networks (CFCN) and dense conditional random fields (CRF). Green depicts correctly predicted liver segmentation, yellow for liver false negative and false positive pixels (all wrong predictions), blue shows correctly predicted lesion segmentation and red lesion false negative and false positive pixels (all wrong predictions). In the first row, the false positive lesion prediction in B of a single UNet as proposed by \cite{Unet} were eliminated in C by CFCN as a result of restricting lesion segmentation to the liver ROI region. In the second row, applying the 3DCRF to CFCN in F increases both liver and lesion segmentation accuracy further, resulting in a lesion Dice score of 82.3\%.}
\label{fig:fig}
\end{figure*}
In the following section, we denote the 3D image volume as $I$, the total number of voxels as $N$ and the set of possible labels as $\labelspace  = \{0,1,\ldots,l\}$. For each voxel $i$, we define a variable $x_i \in \labelspace$ that denotes the assigned label. The probability of a voxel $i$ belonging to label $k$ given the image $I$ is described by $P(x_i=k \vert I)$ and will be modelled by the FCN. 
In our particular study, we use $\labelspace = \{0,1,2\}$ for background, liver and lesion, respectively.

\subsection{3DIRCADb Dataset}
For clinical routine usage, methods and algorithms have to be developed, trained and evaluated on heterogeneous real-life data. Therefore, we evaluated our proposed method on the 3DIRCADb dataset\footnote{The dataset is available on \url{http://ircad.fr/research/3d-ircadb-01}}\cite{soler20123d}. In comparison to the the grand challenge datasets, the 3DIRCADb dataset offers a higher variety and complexity of livers and its lesions and is publicly available. The 3DIRCADb dataset includes 20 venous phase enhanced CT volumes from various European hospitals with different CT scanners. For our study, we trained and evaluated our models using the 15 volumes containing hepatic tumors in the liver with 2-fold cross validation. The analyzed CT volumes differ substantially in the level of contrast-enhancement, size and number of tumor lesions (1 to 42).
We assessed the performance of our proposed method using the quality metrics introduced in the grand challenges for liver and lesion segmentation by \cite{Heimann,deng2008editorial}. 

\subsection{Data preparation, processing and pipeline}
\label{sec:pipeline}
\begin{figure*}[tb]
\centering
\includegraphics[width=\textwidth]{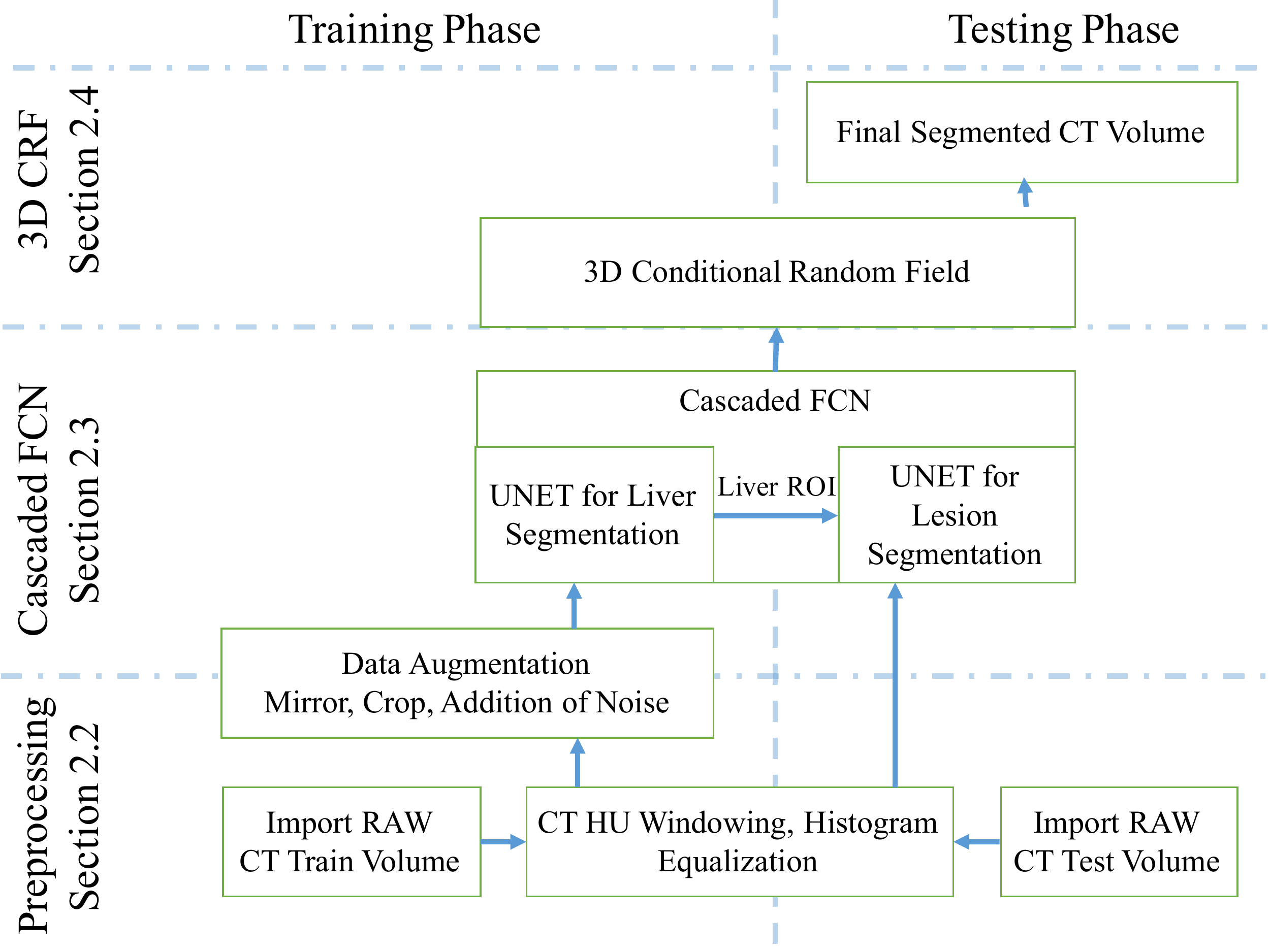}
\caption{Overview of the proposed image segmentation pipeline. In the training phase, the CT volumes are trained after pre-processing and data augmentation in a cascaded fully convolutional neural network (CFCN). To gain the final segmented volume, the test volume is fed-forward in the (CFCN) and refined afterwards using a 3D conditional random field 3DCRF.}
\label{fig:pipeline}
\end{figure*}

Pre-processing was carried out in a slice-wise fashion. First, the Hounsfield unit values were windowed in the range $[-100,400]$ to exclude irrelevant organs and objects, then we increased contrast through histogram equalization. As in \cite{Unet}, to teach the network the desired invariance properties, we augmented the data by applying translation, rotation and addition of gaussian noise. Thereby resulting in an increased training dataset of 22,693 image slices, which were used to train two cascaded FCNs based on the UNet architecture \cite{Unet}. The predicted segmentations are then refined using dense 3D Conditional Random Fields. The entire pipeline is depicted in \autoref{fig:pipeline}.

\subsection{Cascaded Fully Convolutional Neural Networks (CFCN)}
\label{sec:cfcn}

We used the UNet architecture \cite{Unet} to compute the soft label probability maps $P(x_i \vert I)$. The UNet architecture enables accurate pixel-wise prediction by combining spatial and contextual information in a network architecture comprising 19 convolutional layers. In our method, we trained one network to segment the liver in abdomen slices (step 1), and another network to segment the lesions, given an image of the liver (step 2). The segmented liver from step 1 is cropped and resampled to the required input size for the cascaded UNet in step 2, which further segments the lesions.

The motivation behind the cascade approach is that it has been shown that UNets and other forms of CNNs learn a hierarchical representation of the provided data. The stacked layers of convolutional filters are tailored towards the desired classification in a data-driven manner, as opposed to designing hand-crafted features for separation of different tissue types. By cascading two UNets, we ensure that the UNet in step 1 learns filters that are specific for the detection and segmentation of the liver from an overall abdominal CT scan, while the UNet in step 2 arranges a set of filters for separation of lesions from the liver tissue. Furthermore, the liver ROI helps in reducing false positives for lesions. 

A crucial step in training FCNs is appropriate class balancing according to the pixel-wise frequency of each class in the data. In contrast to \cite{long2014fully}, we observed that training the network to segment small structures such as lesions is not possible without class balancing, due to the high class imbalance. Therefore we introduced an additional weighting factor $\omega^{class}$ in the cross entropy loss function $L$ of the FCN.
\begin{equation}
    L =- \frac{1}{n} \sum\limits_{i=1}^N \omega_i^{class}  \left[ \hat{P_i} \log P_i + (1 - \hat{P_i}) \log(1 - P_i) \right]
\end{equation}
$P_i$ denotes the probability of voxel $i$ belonging to the foreground, $\hat{P_i}$ represents the ground truth. We chose $\omega^{class}_i$ to be $\frac{1}{\vert {\text{Pixels of Class } x_i=k} \vert}$.

The CFCNs were trained on a NVIDIA Titan X GPU, using the deep learning framework caffe \cite{jia2014caffe}, at a learning rate of 0.001, a momentum of 0.8 and a weight decay of 0.0005.
\subsection{3D Conditional Random Field (3DCRF)}
\label{sec:3dcrf}
Volumetric FCN implementation with 3D convolutions is strongly limited by GPU hardware and available VRAM \cite{prasoon13}. In addition, the anisotropic resolution of medical volumes (e.g. 0.57-0.8mm in xy and 1.25-4mm in z voxel dimension in 3DIRCADb) complicates the training of discriminative 3D filters. Instead, to capitalise on the locality information across slices within the dataset, we utilize 3D dense conditional random fields CRFs as proposed by \cite{Krahenbuhl2012}. To account for 3D information, we consider all slice-wise predictions of the FCN together in the CRF applied to the entire volume at once.

We formulate the final label assignment given the soft predictions (probability maps)
from the FCN as \emph{maximum a posteriori} (MAP)
inference in a dense CRF, allowing us to consider both
spatial coherence and appearance.

We specify the dense CRF following \cite{Krahenbuhl2012} on the
complete graph $\graph=(\vertices, \edges)$ with vertices $i \in
\vertices$ for each voxel in the image and edges $e_{ij} \in \edges =
\lbrace (i, j) \enspace \forall i, j \in \vertices \enspace
\mathrm{s.t.} \enspace i < j \rbrace$ between \emph{all} vertices. The
variable vector $\x \in \labelspace^N$ describes the label of each
vertex $i \in \vertices$. The energy function that induces the according
Gibbs distribution 
is then given as:

\begin{equation}
  \label{eq:energy}
  E(\x) = \sum_{i \in \vertices} \phi_i(x_i) + \sum_{(i,j) \in \edges} \phi_{ij}(x_i, x_j) \enspace ,
\end{equation}
where $\phi_i(x_i) = -\log \cp{x_i}{I}$ are the unary potentials that
are derived from the FCNs probabilistic output,
$\cp{x_i}{I}$. $\phi_{ij}(x_i,x_j)$ are the pairwise potentials, which
we set to:
\begin{eqnarray}
  \label{eq:pairwisepot}
  \phi_{ij}(x_i, x_j) =  &\mu(x_i, x_j) \bigg( w_{\mathrm{pos}}  \exp  \left( -\frac{\vert p_i - p_j \vert^2}{2 \sigma_{\mathrm{pos}}^2} \right) \qquad \notag \\&+ w_{\mathrm{bil}} \exp \left( -\frac{\vert p_i - p_j \vert^2}{2 \sigma_{\mathrm{bil}}^2} -\frac{\vert I_i - I_j \vert^2}{2 \sigma_{\mathrm{int}}^2}\right) \bigg) \enspace ,
\end{eqnarray}
where $\mu(x_i,x_j) = \mathbf{1}(x_i \neq x_j)$ is the Potts function,
$\vert p_i - p_j \vert$ is the spatial distance between voxels $i$ and
$j$ and $\vert I_i - I_j \vert$ is their intensity difference in the
original image. The influence of the pairwise terms can be adjusted
with their weights $w_{\mathrm{pos}}$ and $w_{\mathrm{bil}}$ and their
effective range is tuned with the kernel widths
$\sigma_{\mathrm{pos}}, \sigma_{\mathrm{bil}}$ and
$\sigma_{\mathrm{int}}$.

We estimate the best labelling $\x^* = \argmin_{\x \in \labelspace^N}
E(\x)$ using the efficient mean field approximation algorithm of \cite{Krahenbuhl2012}. The weights and kernels of the CRF were chosen using a random search algorithm.
\section{Results and Discussion}

The qualitative results of the automatic segmentation are presented in \autoref{fig:fig}. The complex and heterogeneous structure of the liver and all lesions were detected in the shown images. The cascaded FCN approach yielded an enhancement for lesions with respect to segmentation accuracy compared to a single FCN as can be seen in \autoref{fig:fig}. 
In general, we observe significant\footnote{Two-sided paired t-test with p-value $< 4 \cdot 10^{-19}$ } additional improvements for slice-wise Dice overlaps of liver segmentations, from mean Dice $93.1\%$ to $94.3\%$ after applying the 3D dense CRF. 

\begin{table}[]
\centering
\begin{tabular}{@{}llccccc@{}}
\toprule
Approach & VOE & RVD & ASD & MSD & DICE \\
   & {[}\%{]} & {[}\%{]} & {[}mm{]} & {[}mm{]} & {[}\%{]} \\ \midrule
 UNET as in \cite{Unet} & $39$ & $87$  & $19.4$ & $119$  & 72.9 \\
 Cascaded UNET  & 12.8  & -3.3 & 2.3 & 46.7 &  93.1\\
 Cascaded UNET + 3D CRF & 10.7 & -1.4 & 1.5 & 24.0  & 94.3 \\
   &  &  &  &  &  \\
 Li et al. \cite{li2015automatic} (liver-only) & $9.2$ & $-11.2$ & $1.6$ & $28.2$&  \\
 Chartrand et al.  \cite{chartrand2014semi} (semi-automatic) & $6.8$ & $1.7$ & $1.6$ & $24$&  \\
Li et al.  \cite{li2013likelihood} (liver-only) &  &  &  & & 94.5  \\
  \bottomrule
\end{tabular}%
\caption{Quantitative segmentation results of the liver on the 3DIRCADb dataset. Scores are reported as presented in the original papers.}
\label{tab:benchmarking}
\end{table}
Quantitative results of the proposed method are reported in \autoref{tab:benchmarking}.  The CFCN achieves higher scores as the single FCN architecture. Applying the 3D CRF improved the segmentations results of calculated metrics further. The runtime per slice in the CFCN is $2\cdot0.2s=0.4$s without and 0.8s with CRF.

In comparison to state-of-the-art, such as \cite{foruzan2015improved,li2013likelihood,li2015automatic,chartrand2014semi}, we presented a framework, which is capable of a combined segmentation of the liver and its lesion. 
\section{Conclusion}
\label{sec:summary}
Cascaded FCNs and dense 3D CRFs trained on CT volumes are suitable for automatic localization and combined volumetric segmentation of the liver and its lesions. Our proposed method competes with state-of-the-art. We provide our trained models under open-source license allowing fine-tuning for other medical applications in CT data \footnote{Trained models are available at \url{https://github.com/IBBM/Cascaded-FCN}}.
Additionally, we introduced and evaluated dense 3D CRF as a post-processing step for deep learning-based medical image analysis. Furthermore, and in contrast to prior work such as \cite{foruzan2015improved,li2013likelihood,li2015automatic}, our proposed method could be generalized to segment multiple organs in medical data using multiple cascaded FCNs. 
%
%
All in all, heterogeneous CT volumes from different scanners and protocols as present in the 3DIRCADb dataset and in clinical trials can be segmented in under 100s each with the proposed approach. We conclude that CFCNs and dense 3D CRFs are promising tools for automatic analysis of liver and its lesions in clinical routine.